\documentclass{article}
\usepackage{spconf,amsmath,graphicx}
\usepackage{times}
\usepackage{epsfig}
\usepackage{graphicx}
\usepackage{amsmath}
\usepackage{amssymb}
\usepackage{verbatim}
\usepackage{algpseudocode}
\usepackage{algorithm}
\usepackage{lineno}
\usepackage[pagebackref=true,breaklinks=true,letterpaper=true,colorlinks,bookmarks=false]{hyperref}

\DeclareMathOperator*{\argmin}{arg\,min}

\title{Hierarchical Sparse and Collaborative Low-Rank Representation for Emotion Recognition}
\name{Xiang Xiang, Minh Dao, Gregory D. Hager, Trac D. Tran
}
\address{Johns Hopkins University,  3400 N. Charles Street, Baltimore, MD 21218, USA \\
\{xxiang, minh.dao, ghager1, trac\}@jhu.edu}

\begin{document}
%
\maketitle

\begin{abstract}
In this paper, we design a Collaborative-Hierarchical Sparse and Low-Rank (C-HiSLR) model that is natural for recognizing human emotion in visual data.
Previous attempts require explicit expression components, which are often unavailable and difficult to recover.
Instead, our model exploits the low-rank property to subtract neutral faces from expressive facial frames
as well as performs sparse representation on the expression components with group sparsity enforced.
For the CK+ dataset, C-HiSLR on raw expressive faces performs as competitive as
the Sparse Representation based Classification (SRC) applied on manually prepared emotions.
Our C-HiSLR performs even better than SRC in terms of true positive rate.
\end{abstract}
\begin{keywords}
Low-rank, group sparsity, multichannel
\end{keywords}
%
\vspace{-2mm}
\section{Introduction} \label{sec:intro}
\begin{figure}
\begin{center}
   \includegraphics[width=1\linewidth]{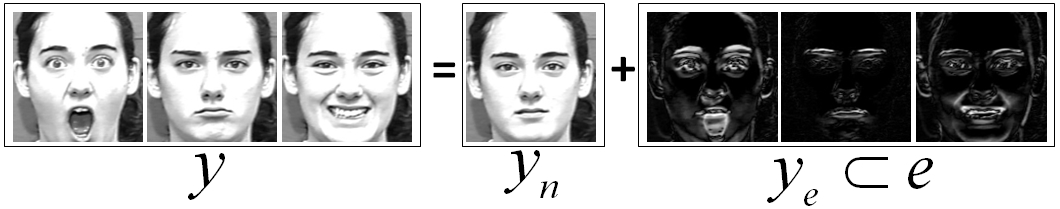}
\end{center}
   \vspace{-8mm}
   \caption{The separability of the neutral face $\mathbf{y}_n$ and emotion $\mathbf{y}_e$. Given a different expressive face $\mathbf{y}$ (\emph{e.g., surprise, sadness, happiness}), the difference is $\mathbf{y}_e$, which is encoded in error $\mathbf{e}$.
}
\label{fig:separate}
\end{figure}

\vspace{-2mm}
In this paper, the problem of interest is to recognize the emotion given a video of a human face and emotion category \cite{exp-survey}.
As shown in Fig.\ref{fig:separate},
an expressive face can be separated into a dominant neutral face and a sparse {\bf expression component},
which we term {\bf emotion} and is usually encoded in a sparse noise term $\mathbf{e}$.
We investigate \emph{if we can sparsely represent the emotion over a dictionary of emotions} \cite{petrou10}
rather than expressive faces \cite{ptucha11},
which may confuse a similar expression with a similar identity \cite{petrou10}.
Firstly, \emph{how to get rid of the neutral face?}
Surely we can prepare an expression with a neutral face explicitly provided as suggested in \cite{petrou10}.
Differently, we treat an emotion as an action and assume neutral faces stay the same.
If we stack vectors of neutral faces as a matrix,
it should be low-rank (ideally with rank 1).
Similarly, over time sparse vectors of emotions form a sparse matrix.
Secondly, \emph{how to recover the low-rank and sparse components?}
In \cite{Taheri11},
the (low-rank) Principal Component Pursuit (PCP)  \cite{rpca} is performed explicitly.
While theoretically the recovery is exact under conditions \cite{rpca},
it is of approximate nature in practice.
Finally, since we only care about the sparse component,
\emph{can we avoid such an approximate explicit PCP step?}
This drives us to exploit Sparse representation and Low-Rank property jointly in one model named SLR (Sec. \ref{sec:SLR}).

Different from image-based methods \cite{petrou10,Taheri11},
we treat an emotion video as a multichannel signal.
If we just use a single channel such as one frame to represent an emotion,
much information is lost since all frames collaboratively represent an emotion.
Therefore, we prefer using all or most of them.
\emph{Should we treat them separately or simultaneously?}
The former just needs to recover the sparse coefficient vector for each frame.
The latter gives a spatial-temporal representation, while it requires the recovery of a sparse coefficient matrix, which should often exhibit a specific structure.
\emph{Should we enforce a class-wise sparsity separately or enforce a group sparsity collaboratively?}
\cite{Taheri11} models the class-wise sparsity separately for the recognition of
a neutral face's identity and an expression image's emotion once they have been separated.
Alternatively, we can exploit the low-rankness as well as structured sparsity by inter-channel observation.
Since class decisions may be inconsistent,
we prefer a collaborative model \cite{eldar10} with group sparsity enforced \cite{Huang10}.
This motivates us to introduce the group sparsity as a root-level sparsity to the SLR model embedded with a leaf-level atom-wise sparsity.
The reason of keeping both levels is that signals over frames share class-wise yet not necessarily attom-wise sparsity patterns \cite{chilasso}.
Therefore, we term this model Collaborative-Hierarchical Sparse and Low-Rank (C-HiSLR) model.

In the remainder of this paper, we review sparse and low-rank representation literature in \mbox{Sec. \ref{sec:bg}},
elaborate our model in \mbox{Sec. \ref{sec:model}}, discuss the optimization in \mbox{Sec. \ref{sec:optm}},
empirically validate the model in \mbox{Sec. \ref{sec:exp}}, and draw a conclusion in \mbox{Sec. \ref{sec:conclu}}.
\vspace{-4mm}

\section{Related Works} \label{sec:bg}
\vspace{-3mm}
When observing a random signal $\mathbf{y}$ for recognition, we hope to send the classifier a \emph{discriminative compact} representation $\mathbf{x}$,
which satisfies $\mathbf{Ax=y}$ and is yet computed by pursuing the best \emph{reconstruction}.
When $\mathbf{A}$ is under-complete,
a closed-form approximate solution can be obtained by
Least-Squares: \\
\indent \indent $\mathbf{x}^*=\argmin_{\mathbf{x}} \|\mathbf{y-Ax}\|_2^2 \approx (\mathbf{ A}^T \mathbf{A})^{-1} \mathbf{A}^T \mathbf{y}$. \\
When $\mathbf{A}$ is over-complete,
we add a Tikhonov regularizer \cite{tik-reg}:
\mbox{$\mathbf{x}^* = \argmin_{\mathbf{x}} \|\mathbf{y-Ax}\|_2^2+\lambda_r \|\mathbf{x}\|_2^2
= \argmin_{\mathbf{x}} \|\mathbf{\widetilde{y}-\widetilde{A}x}\|_2^2$} \\
$\approx (\mathbf{A}^T \mathbf{A}+\lambda_r \mathbf{I})^{-1} \mathbf{A}^T \mathbf{y}$
where $\mathbf{\widetilde{y}=[y,0]}^T$; $\mathbf{\widetilde{A}=[A},\sqrt{\lambda_r}\mathbf{I]}^T$
is always under-complete.
But $\mathbf{x}^*$ is not necessarily compact yet generally dense.
Alternatively, we can seek a sparse usage of $\mathbf{A}$.
Sparse Representation based Classification \cite{SRC} (SRC) expresses a test sample $\mathbf{y}$
as a linear combination $\mathbf{y=Dx}$ of \emph{training samples} stacked columnwise in a dictionary $\mathbf{D}$.
\emph{Since non-zero coefficients should all drop to the ground-truth class,
ideally not only $\mathbf{x}$ is sparse but also the class-level sparsity is $1$.}
In fact, non-zero coefficients also drop to other classes due to noises and correlations among classes.
By adding a sparse error term $\mathbf{e}$, SRC simply employs an atom-wise sparsity: \\
\indent \indent \indent $[\mathbf{x}^*,\mathbf{e}^*]^T = \argmin_{\widetilde{\mathbf{x}}}  sparsity(\widetilde{\mathbf{x}})$\\
\indent \emph{s.t.} \indent $\mathbf{y=Dx+e}=[\hspace{0.2em} \mathbf{D} \hspace{0.3em}|\hspace{0.3em} \mathbf{I} \hspace{0.2em}] \times \left[ \begin{array}{c} \mathbf{x} \\ \mathbf{e}\end{array} \right] = \widetilde{\mathbf{D}}\widetilde{\mathbf{x}}$, \\
where $\widetilde{\mathbf{D}}$ is over-complete and needs to be sparsely used.
SRC evaluates which class leads to the minimum reconstruction error,
which can be seen as a max-margin classifier \cite{margin-src}.
Using a fixed $\mathbf{D}$ without dictionary learning \cite{Suo14} or sparse coding,
SRC still performs robustly well for denoising and coding tasks such as well-aligned noisy face identifications.
But there is a lack of theoretical justification why a sparser representation is more discriminative.
\cite{huang06nips} incorporates the Fisher's discrimination power into the objective.
\cite{CRC} follows the regularized Least-Squares \cite{tik-reg}
and argues SRC's success is due to the linear combination
as long as the ground-truth class dominates coefficient magnitudes.
SRC's authors clarify this confusion using more tests on robustness to noises \cite{yimaclarify}.

In practice, we care more about how to recover $\mathbf{x}$ \cite{OMP}.
Enforcing sparsity is feasible since $\mathbf{x}$ can be exactly recovered from $\mathbf{y=Dx+e}$ under conditions for $\mathbf{D}$ \cite{exrec}.
However, finding the sparsest solution is
NP-hard and difficult to solve exactly \cite{sparse-nphard}.
But now, it is well-known that the $\ell_1$ norm is a good convex relaxation of sparsity
\--\--
minimizing the $\ell_1$ norm induces the sparsest solution under mild conditions \cite{l0l1eq}.
Exact recovery is also guaranteed by $\ell_1$-minimization under suitable conditions \cite{l1-exact}.
Typically, an iterative greedy algorithm is the Orthogonal Matching Pursuit (OMP) \cite{OMP}.

For multichannel $\mathbf{Y}$ with dependant coefficients across channels \cite{lrank}, $\mathbf{Y=DX}$ where $\mathbf{X}$ is low-rank.
In a \emph{unsupervised} manner, Sparse Subspace Clustering \cite{SSC} of $\mathbf{Y}$ solves $\mathbf{Y=YX}$ where $\mathbf{X}$ is sparse and Principal Component Analysis is $\min_{\mathbf{A}} \|\mathbf{Y-AY}\|^2$ where $\mathbf{A}$ is a projection matrix.



   \vspace{-3mm}
\section{Representation Models} \label{sec:model}
   \vspace{-2mm}
In this section, we explain how to model $\mathbf{X}$ using $\mathbf{Y}$ and training data $\mathbf{D}$,
which contains $K \in \mathbb{Z^+}$ types of {\bf emotions}.
We would like to classify a test video as one of the $K$ classes.
   \vspace{-8mm}
\subsection{SLR: joint Sparse representation and Low-Rankness} \label{sec:SLR}
\vspace{-2mm}
First of all, we need an explicit representation $\mathbf{Y}$ of an expressive face.
The matrix $\mathbf{Y} \in \mathbb{R}^{d \times \tau}$ can be an arrangement
of $d$-dimensional feature vectors
$\mathbf{y} \in \mathbb{R}^{d}$ ($i=1,2,...,\tau$) such as Gabor features \cite{gabor_src}
or concatenated image raw intensities \cite{SRC} of the $\tau$ frames:
$\mathbf{Y} = \big[\mathbf{Y}_1|...|\mathbf{Y}_{\tau} \big]_{d \times \tau}$.
We emphasize our model's power by simply using the raw pixel intensities.

\begin{figure}
\begin{center}
   \includegraphics[width=0.97\linewidth]{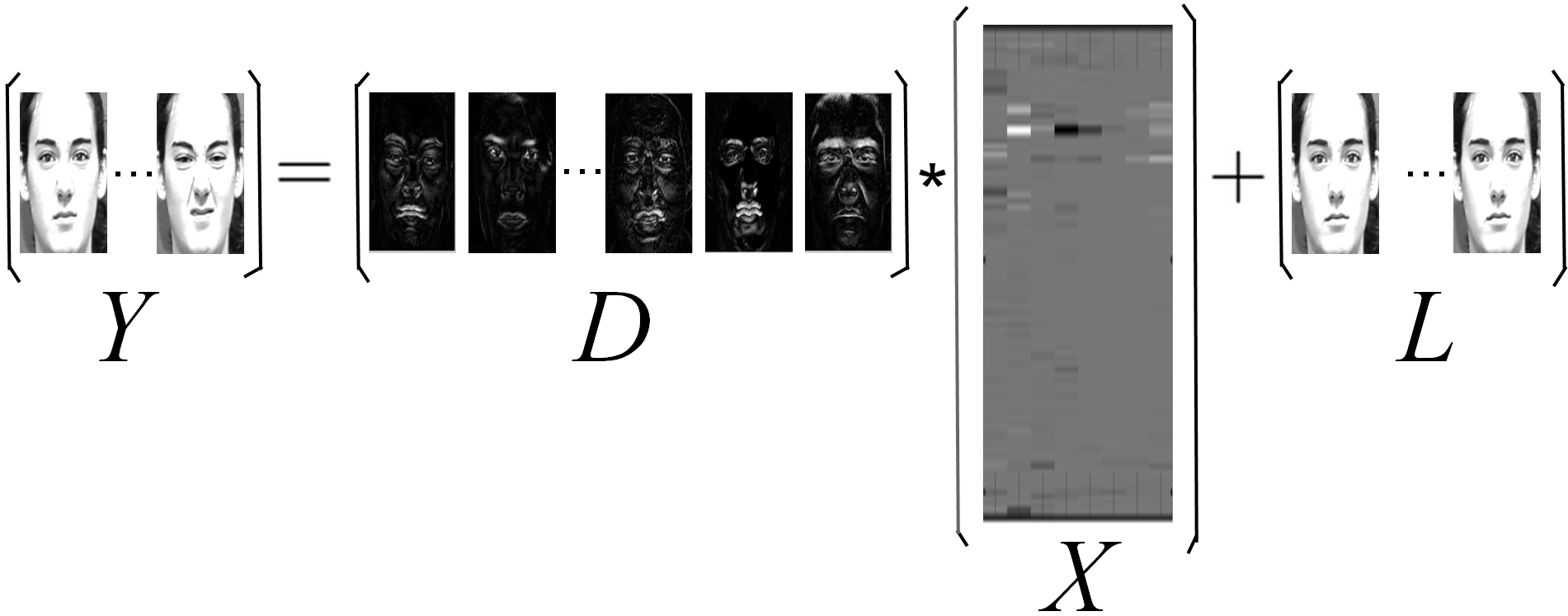}
\end{center}
   \vspace{-8mm}
   \caption{Pictorial illustration of the constraint in SLR and C-HiSLR for recognizing \emph{disgust}. $\mathbf{D}$ is prepared and fixed. }
\label{fig:constraint}
\end{figure}

Now, we seek an implicit latent representation $\mathbf{X} \in \mathbb{R}^{n \times \tau}$ of an input test face's emotion $\mathbf{ Y}_e \in \mathbb{R}^{d \times \tau}$
as a sparse linear combination of prepared fixed training emotions $\mathbf{D} \in \mathbb{R}^{d \times n}$:\\
\indent \indent \indent \indent \indent \indent
$\mathbf{Y}_e = \mathbf{DX}$. \\
Since an expressive face $\mathbf{y}=\mathbf{y}_e+\mathbf{y}_n$ is a superposition of an emotion $\mathbf{y}_e \in \mathbb{R}^d$ and a neutral face $\mathbf{y}_n \in \mathbb{R}^d$, we have \\
\indent \indent \indent \indent \indent \indent
$\mathbf{Y} = \mathbf{Y}_e + \mathbf{L}$, \\
where $\mathbf{L} \in \mathbb{R}^{d \times \tau}$ is ideally $\tau$-times repetition of the column vector of a neutral face $\mathbf{y}_n \in \mathbb{R}^d$. Presumably $\mathbf{L} = \big[ \mathbf{y}_n|...|\mathbf{y}_n \big]_{d \times \tau}$. As shown in Fig. \ref{fig:constraint}, $\mathbf{X}$ subjects to \\
\indent \indent \indent \indent \indent \indent
$\mathbf{Y = DX + L}$, \\
where the dictionary matrix $\mathbf{D}_{d \times n}$ is an arrangement of \mbox{all}
sub-matrices $\mathbf{ D}_{[j]}$, $j=1,...,\lfloor \frac{n}{\tau} \rfloor$.
\emph{Only for training}, we have $\lfloor \frac{n}{\tau} \rfloor$ training emotions with neutral faces subtracted.
The above constraint of $\mathbf{X}$ characterizes an affine transformation from the latent representation $\mathbf{X}$ to the observation $\mathbf{Y}$.
If we write $\mathbf{X}$ and $\mathbf{Y}$ in homogeneous forms \cite{HomoCoord}, then we have \\
\indent \indent $\left[ \begin{array}{c} \mathbf{Y}_{d \times \tau} \\ \mathbf{1}_{1 \times \tau} \end{array} \right]
= \begin{bmatrix} \mathbf{D}_{d \times n} & \big(\mathbf{y}_n\big)_{d \times 1} \\ \mathbf{0}_{1 \times n} & 1 \end{bmatrix} \times \left[ \begin{array}{c} \mathbf{X}_{n \times \tau} \\ \mathbf{1}_{1 \times \tau} \end{array} \right]$. \\
In the ideal case with $rank(\mathbf{L})=1$,
if the neutral face $\mathbf{y}_n$ is pre-obtained \cite{petrou10,Taheri11},
it is trival to solve for $\mathbf{X}$.
Normally, $\mathbf{y}_n$ is unknown and $\mathbf{L}$ is not with rank $1$ due to noises.
As $\mathbf{X}$ is supposed to be sparse and $rank(\mathbf{L})$ is expected to be as small as possible (maybe even $1$),
intuitively our objective is to \\
\indent \indent \indent
$\min_{\mathbf{X,L}} sparsity(\mathbf{X}) + \lambda_L \cdot rank(\mathbf{L})$, \\
where
$rank(\mathbf{L})$ can be seen as the sparsity of the vector formed by the singular values of $\mathbf{L}$.
Here $\lambda_L$ is a non-negative weighting parameter we need to tune \cite{gsure}.
When $\lambda_L = 0$, the optimization problem reduces to that in SRC.
With both terms relaxed to be $\ell_1$ norm, we alternatively solve
\indent \indent \indent \indent \indent
$\min_{\mathbf{X,L}} \| \mathbf{X} \|_1 + \lambda_L \| \mathbf{L} \|_*$,  \\
where $\| \cdot \|_1$ is the entry-wise $\ell_1$ matrix norm,
whereas $\| \cdot \|_*$ is the Schatten $\ell_1$ matrix norm (nuclear norm, trace norm) which can be seen as applying $\ell_1$ norm to the vector of singular values. Now, the proposed joint SLR model is expressed as \\
   \vspace{-2mm}
\begin{equation} \label{eq:slr}
\begin{aligned}
\min_{\mathbf{X,L}} \| \mathbf{X} \|_1 + \lambda_L \| \mathbf{L} \|_*  \indent s.t. \indent \mathbf{Y = DX + L}
\end{aligned}
\end{equation}

\noindent We solve (\ref{eq:slr}) for matrices $\mathbf{X}$ and $\mathbf{L}$ by the Alternating Direction Method of Multipliers (ADMM) \cite{ADMM} (see Sec. \ref{sec:optm}).
   \vspace{-3mm}
\subsection{C-HiSLR: a Collaborative-Hierarchical SLR model} \label{sec:hslr}
   \vspace{-2mm}
If there is no low-rank term $\mathbf{L}$, (\ref{eq:slr}) becomes a problem of multi-channel Lasso (Least Absolute Shrinkage and Selection Operator).
For a single-channel signal, Group Lasso \cite{GLasso} has explored the group structure for Lasso yet does not enforce sparsity within a group,
while Sparse Group Lasso \cite{SGLasso} yields an atom-wise sparsity as well as a group sparsity.
Then, \cite{chilasso} extends Sparse Group Lasso to multichannel,
resulting in a Collaborative-Hierarchical Lasso (C-HiLasso) model.
For our problem, we do need $\mathbf{L}$, which induces a Collaborative-Hierarchical Sparse and Low-Rank (C-HiSLR) model:
   \vspace{-2mm}
\begin{equation} \label{eq:hslr}
\begin{aligned}
\min_{\mathbf{X,L}} \| \mathbf{X} \|_1 & + \lambda_L \| \mathbf{L} \|_* + \lambda_g \sum_{G \in \mathcal{G}} {\| \mathbf{X}^{[G]} \|}_F \\
& s.t. \indent \mathbf{Y = DX + L}
\end{aligned}
\end{equation}
   \vspace{-3mm}
\begin{figure}
\begin{center}
   \includegraphics[width=0.72\linewidth]{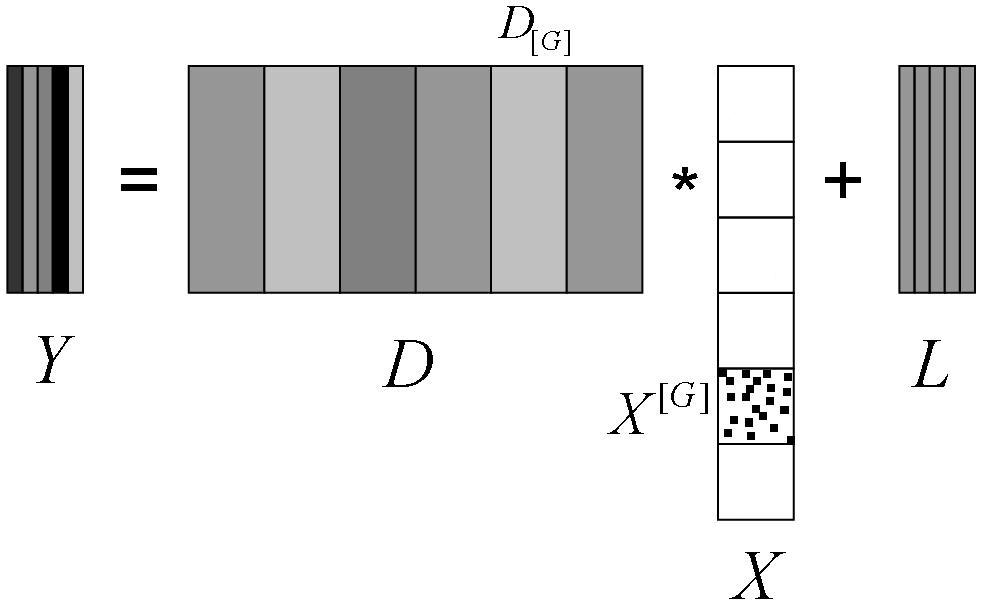}
\end{center}
   \vspace{-8mm}
   \caption{Pictorial illustration of the constraint in the C-HiSLR.}
\label{fig:gspa}
\end{figure}

\noindent where $\mathbf{X}^{[G]}$ is the sub-matrix formed by all the {\bf rows} indexed by the elements in group $G \subseteq \{1,...,$n$\}$.
As shown in \mbox{Fig. \ref{fig:gspa}},
given a group $G$ of indices, the sub-dictionary of {\bf columns} indexed by $G$ is denoted as $\mathbf{D}_{[G]}$.
$\mathcal{G} = \{G_1,...,G_K\}$ is a non-overlapping partition of $\{1,...,n\}$.
Here ${\| \cdot \|}_F$ denotes the Frobenius norm, which is the entry-wise $\ell_2$ norm as well as the Schatten $\ell_2$ matrix norm and can be seen as a group's magnitude.
$\lambda_g$ is a non-negative weighting parameter for the group regularizer,
which is generalized from an $\ell_1$ regularizer (consider $\mathcal{G}=\big\{\{1\},\{2\},...,\{n\}\big\}$ for singleton groups) \cite{chilasso}.
When $\lambda_g = 0$, C-HiSLR degenerates into SLR.
When $\lambda_L = 0$, we get back to collaborative Sparse Group Lasso.
   \vspace{-3mm}
\subsection{Classification} \label{sec:cls}
   \vspace{-2mm}
Following SRC, for each class $c \in \{1,2,...,K\}$, let $\mathbf{D}_{[G_c]}$ denote the sub-matrix of $\mathbf{D}$ which consists of all the columns of $\mathbf{D}$ that correspond to emotion class $c$
and similarly for $\mathbf{X}^{[G_c]}$.
We classify $\mathbf{Y}$ by assigning it to the class with minimal residual as
  \mbox{$c^* = \argmin_c r_c(\mathbf{Y}) := ||\mathbf{Y} - \mathbf{D}_{[G_c]}\mathbf{X}^{[G_c]} - \mathbf{L}||_F$}.

   \vspace{-1mm}
\section{Optimization} \label{sec:optm}
   \vspace{-3mm}
Both SLR and C-HiSLR models can be seen as solving
   \vspace{-2mm}
\begin{equation} \label{eq:gene-slr}
\begin{aligned}
\min_{\mathbf{X,L}} f(\mathbf{X}) + \lambda_L \| \mathbf{L} \|_* \indent s.t. \indent \mathbf{Y = DX + L}
\end{aligned}
\vspace{-2mm}
\end{equation}

To follow a standard iterative ADMM procedure \cite{ADMM},
we write down the augmented Lagrangian function for \eqref{eq:gene-slr} as
\vspace{-2mm}
\begin{equation}\label{al}
\begin{split}
&  \mathcal{L}(\mathbf{X},\mathbf{L},\mathbf{\Lambda}) = f(\mathbf{X}) + \lambda_L ||\mathbf{L}||_* \\
&  + \left\langle \mathbf{\Lambda}, \mathbf{Y-DX-L}\right\rangle + \frac{\beta}{2}||\mathbf{Y-DX-L}||_F^2,
\end{split}
\vspace{-5mm}
\end{equation}
where $\mathbf{\Lambda}$ is the matrix of multipliers,
$\left\langle \cdot,\cdot \right\rangle$ is inner product,
and $\beta$ is a positive weighting parameter for the penalty (augmentation).
A single update at the $k$-th iteration includes
{\footnotesize
\begin{align}
  \mathbf{L}_{k+1} &= \argmin_{\mathbf{L}} \lambda_L||\mathbf{L}||_* + \frac{\beta}{2}||\mathbf{Y}-\mathbf{D}\mathbf{X}_k-\mathbf{L}+\frac{1}{\beta}\mathbf{\Lambda}_k||_F^2 \label{LStep} \\
	\mathbf{X}_{k+1} &= \argmin_{\mathbf{X}} f(\mathbf{X}) + \frac{\beta}{2}||\mathbf{Y-DX} -\mathbf{L}_{k+1}+\frac{1}{\beta}\mathbf{\Lambda}_k||_F^2 \label{XStep} \\
	\mathbf{\Lambda}_{k+1} &= \mathbf{\Lambda}_k + \beta(\mathbf{Y}-\mathbf{D}\mathbf{X}_{k+1}-\mathbf{L}_{k+1}). \label{MultStep}
\end{align}}
The sub-step of solving \eqref{LStep} has a closed-form solution:
   \vspace{-3mm}
\begin{equation}\label{LSol}
  \mathbf{L}_{k+1} = \mathcal{D}_\frac{\lambda_L}{\beta}(\mathbf{Y}-\mathbf{DX}_k+\frac{1}{\beta}\mathbf{\Lambda}_k),
  \vspace{-3mm}
\end{equation}
where $\mathcal{D}$ is the shrinkage thresholding operator.
In SLR where $f(\mathbf{X}) = \| \mathbf{X} \|_1$, \eqref{XStep} is a Lasso problem, which we solve by using an existing fast solver \cite{yang08l1}. When $f(\mathbf{X})$ follows \eqref{eq:hslr} of C-HiSLR, computing $\mathbf{X}_{k+1}$ needs an approximation based on the Taylor expansion at $\mathbf{X}_{k}$ \cite{Minh-rank,chilasso}.
We refer the reader to \cite{chilasso} for the convergence analysis and recovery guarantee.
   \vspace{-5mm}
\section{Experimental Results} \label{sec:exp}
\vspace{-2mm}

\begin{figure}
\begin{center}
   \includegraphics[width=0.8\linewidth]{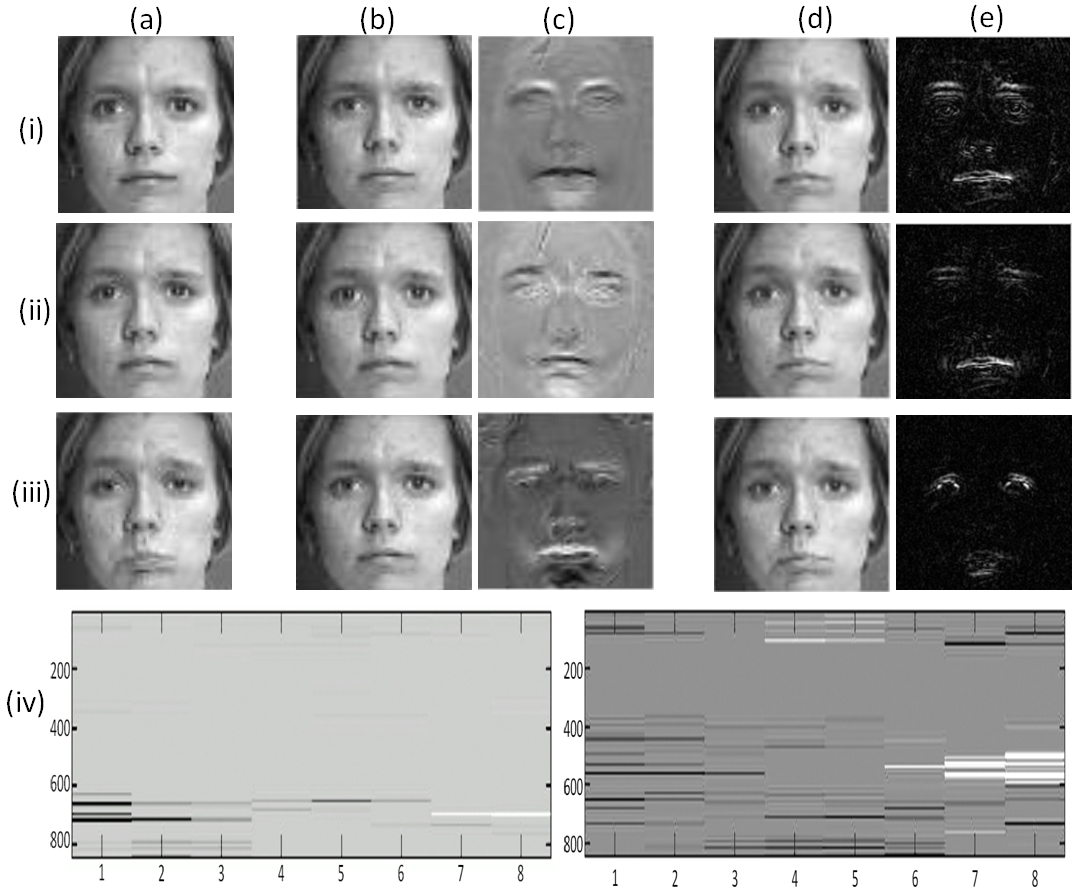}
\end{center}
\vspace{-8mm}
    \caption{Effect of group sparsity. $\tau_{trn}=8$. (a) is the test input $\mathbf{Y}$.
    (b)(c) are recovered $\mathbf{L}$ and $\mathbf{DX}$, given by \mbox{{\bf C-HiSLR}} which correctly classifies (a) as \emph{contempt}.
    (d)(e) are recovery results given by {\bf SLR} which mis-classifies (a) as sadness.
    (i),(ii),(iii) denote results of frame \#1, \#4, \#8 respectively,
    whereas (iv) displays the recovered $\mathbf{X}$
    (left for C-HiSLR and right for SLR).
    $\mathbf{X}$ given by C-HiSLR is group-sparse as we expected.
    }
\label{fig:chi-recovery}
\end{figure}

\begin{figure}
\begin{center}
   \includegraphics[width=0.8\linewidth]{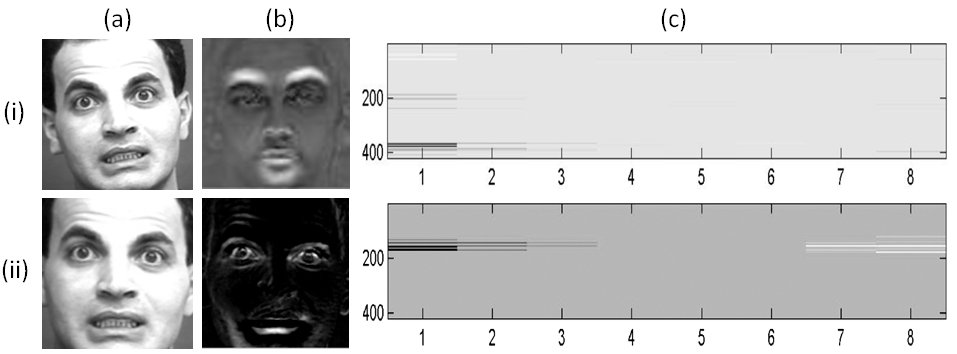}
\end{center}
\vspace{-8mm}
    \caption{Effect of alignment. Shown for {\bf C-HiSLR} with $\tau_{trn}=4$. (a) is the test input (\emph{fear}). (b) and (c) are recovered $\mathbf{DX}$ and $\mathbf{X}$, respectively.
    (i) is under {\bf imperfect} alignment while (ii) is under {\bf perfect} alignment.
    $\mathbf{X}$ in (i) is \emph{not} group-sparse. }
\label{fig:align}
\end{figure}

All experiments are conducted on the CK+ dataset \cite{ck+} which consists of 321 emotion sequences with labels (angry, contempt\footnote{Contempt is discarded in \cite{petrou10,Taheri11} due to its confusion with other classes.}, disgust, fear, happiness, sadness, surprise) \footnote{Please visit \url{http://www.cs.jhu.edu/~xxiang/slr/} for the cropped face data and programs of C-HiSLR, SLR, SRC and Eigenface.}
and is randomly divided into a training set (10 sequences per category) and a testing set (5 sequences per category).
\emph{For SRC, we assume that the information of neutral face is provided.}
We subtract the first frame (a neutral face) from the last frame per sequence for both training and testing.
Thus, each emotion is represented as an image.
However, \emph{for SLR and C-HiSLR, we assume no prior knowledge of the neutral face.}
We form a dictionary by subtracting the first frame from the last $\tau_{trn}$ frames per sequence
and form a testing unit using the last ($\tau_{tst}-1$) frames together with the first frame, which is {\bf not} explicitly known as a neutral face.
Thus, each emotion is represented as a video.
Here, we set $\tau_{trn}=4$ or $8$, $\tau_{tst}=8$, $\lambda_{L}=10$ and $\lambda_{G}=4.5$.
\mbox{Fig. \ref{fig:chi-recovery}} visualizes the recovery results given by C-HiSLR.
Facial images are cropped using the Viola-Jones detector \cite{cvtbox} and resized to $64 \times 64$.
As shown in \mbox{Fig. \ref{fig:align}}, imperfect alignment may affect the performance.

Firstly, SRC achieves a total recognition rate of {\bf 0.80},
against {\bf 0.80} for eigenface with nearest subspace classifier
and {\bf 0.72} for eigenface with nearest neighbor classifier.
This verifies that emotion is sparsely representable by training data
and SRC can be an alternative to subspace based methods.
Secondly, Table \ref{tb-gspa}-\ref{tb-src-omp} present the confusion matrix ($\tau_{trn}=4$)
and Table \ref{tb-sum} summarizes the true positive rate (\emph{i.e.}, sensitivity).
We have anticipated that SLR ({\bf 0.70}) performs worse than SRC ({\bf 0.80}) since SRC is equipped with neutral faces.
However, C-HiSLR's result ({\bf 0.80}) is comparable with SRC's. C-HiSLR performs even better in terms of sensitivity,
which verifies that the group sparsity indeed boosts the performance.

\vspace{-4mm}
\section{Conclusion} \label{sec:conclu}
\vspace{-3mm}
We design the C-HiSLR representation model for emotion recognition,
unlike \cite{petrou10} requiring neutral faces as inputs
and \cite{Taheri11} generating labels of identity and emotion as mutual by-products with extra efforts.
Our contribution is two-fold.
First, we do not recover \mbox{emotion} explicitly.
Instead, we treat frames simultaneously and implicitly subtract the low-rank neutral face.
Second, we preserve the label consistency by enforcing atom-wise as well as group sparsity.
For the CK+ dataset, C-HiSLR's performance on raw data
is comparable with SRC given neutral faces,
which verifies that emotion is automatically separable from expressive faces as well as sparsely representable.
Future works will include handling misalignment \cite{hager98} and incorporating dictionary learning \cite{Suo14}.  \\

\indent \indent \indent \indent \indent {\bf ACKNOWLEDGMENTS} \\
This work is supported by US National Science Foundation
under Grants CCF-1117545 and CCF-1422995, Army Research Office
under Grant 60219-MA, and Office of Naval Research under Grant
N00014-12-1-0765.
The first author is grateful for the fellowship from China Scholarship Council.

\vspace{-10mm}
\begin{table}
{\small
\begin{center}
\begin{tabular}{|c|c|c|c|c|c|c|c|}
\hline
& An & Co & Di & Fe & Ha & Sa & Su \\
\hline
\hline
An & \bf{0.77} & 0.01 & 0.09 & 0.02 & 0 & 0.07 & 0.04 \\
\hline
Co & 0.08 & \bf{0.84} & 0 & 0 & 0.03 & 0.04 & 0 \\
\hline
Di & 0.05 & 0 & \bf{0.93} & 0.01 & 0.01 & 0.01 & 0 \\
\hline
Fe & 0.09 & 0.01 & 0.03 & \bf{0.53} & 0.12 & 0.07 & 0.15 \\
\hline
Ha & 0.01 & 0.02 & 0.01 & 0.02 & \bf{0.93} & 0 & 0.03 \\
\hline
Sa & 0.19 & 0.02 & 0.02 & 0.05 & 0 & \bf{0.65} & 0.07 \\
\hline
Su & 0 & 0.02 & 0 & 0.02 & 0 & 0.02 & \bf{0.95} \\
\hline
\end{tabular}
\end{center}
\vspace{-6mm}
\caption{Confusion matrix for {\bf C-HiSLR} on CK+ dataset \cite{ck+} without explicitly knowing neutral faces. Columns are predictions and rows are ground truths. We randomly choose 15 sequences for training and 10 sequences for testing per class. We let the optimizer run for 600 iterations. Results are averaged over 20 runs and rounded to the nearest. The total recognition rate is {\bf 0.80} with a standard deviation of 0.05. }
\label{tb-gspa}
}
\end{table}
   \vspace{-2mm}
\begin{table}
{\small
\begin{center}
\begin{tabular}{|c|c|c|c|c|c|c|c|}
\hline
& An & Co & Di & Fe & Ha & Sa & Su \\
\hline
\hline
An & \bf{0.51} & 0 & 0.10 & 0.02 & 0 & 0.31 & 0.06 \\
\hline
Co & 0.03 & \bf{0.63} & 0.03 & 0 & 0.04 & 0.26 & 0.01 \\
\hline
Di & 0.04 & 0 & \bf{0.74} & 0.02 & 0.01 & 0.15 & 0.04 \\
\hline
Fe & 0.08 & 0 & 0.01 & \bf{0.51} & 0.03 & 0.19 & 0.18 \\
\hline
Ha & 0 & 0.01 & 0 & 0.03 & \bf{0.85} & 0.08 & 0.03 \\
\hline
Sa & 0.09 & 0 & 0.04 & 0.04 & 0 & \bf{0.70} & 0.13 \\
\hline
Su & 0 & 0.01 & 0 & 0.02 & 0.01 & 0.02 & \bf{0.94} \\
\hline
\end{tabular}
\end{center}
\vspace{-6mm}
\caption{Confusion matrix for {\bf SLR} on CK+ dataset without explicit neutral faces. We randomly choose 15 sequences for training and 10 for testing per class.
We let the optimizer run for 100 iterations and Lasso run for 100 iterations.
Results are averaged over 20 runs and rounded to the nearest. The total recognition rate is {\bf 0.70} with a standard deviation of 0.14. }
\label{tb-jslr}
}
\end{table}

\begin{table}
{\small
\begin{center}
\begin{tabular}{|c|c|c|c|c|c|c|c|}
\hline
& An & Co & Di & Fe & Ha & Sa & Su \\
\hline
\hline
An & \bf{0.71} & 0.01 & 0.07 & 0.02 & 0.01 & 0.03 & 0.16 \\
\hline
Co & 0.07 & \bf{0.60} & 0.02 & 0 & 0.16 & 0.03 & 0.12 \\
\hline
Di & 0.04 & 0 & \bf{0.93} & 0.02 & 0.01 & 0 & 0 \\
\hline
Fe & 0.16 & 0 & 0.09 & \bf{0.25} & 0.25 & 0 & 0.26 \\
\hline
Ha & 0.01 & 0 & 0 & 0.01 & \bf{0.96} & 0 & 0.02 \\
\hline
Sa & 0.22 & 0 & 0.13 & 0.01 & 0.04 & \bf{0.24} & 0.35 \\
\hline
Su & 0 & 0.01 & 0 & 0 & 0.01 & 0 & \bf{0.98} \\
\hline
\end{tabular}
\end{center}
\vspace{-6mm}
\caption{Confusion matrix for {\bf SRC} \cite{SRC} with neutral faces explicitly provided,
in a similar setting with \cite{petrou10}.
We choose half of the dataset for training and the other half for testing per class.
The optimizer is OMP and the sparsity level is set to 35\%.
Results are averaged over 20 runs and rounded to the nearest. The total recognition rate is {\bf 0.80} with a standard deviation of 0.05. The rate for \emph{fear} and \emph{sad} are especially low.  }
\label{tb-src-omp}
}
\end{table}

\begin{table}
{\small
\begin{center}
\begin{tabular}{|c|c|c|c|c|c|c|c|}
\hline
Model   & An & Co & Di & Fe & Ha & Sa & Su \\
\hline
    SRC & 0.71 & \emph{0.60} & {\bf 0.93} & \emph{0.25} & {\bf 0.96} & \emph{0.24} & {\bf 0.98} \\
\hline
    SLR & \emph{0.51} & \emph{0.63} & \emph{0.74} & {\bf 0.51} & \emph{0.85} & {\bf 0.70} & {\bf \emph{0.94}} \\
\hline
C-HiSLR & {\bf 0.77} & {\bf 0.84} & {\bf 0.93} & {\bf 0.53} & {\bf 0.93} & {\bf 0.65} & {\bf \emph{0.95}} \\
\hline
\end{tabular}
\end{center}
\vspace{-6mm}
\caption{Comparison of sensitivity. The {\bf bold} and \emph{italics} denote the highest and lowest respectively. Difference within $0.05$ is treated as comparable. C-HiSLR performs the best.}
\label{tb-sum}
}
\end{table}

\vfill\pagebreak

{\footnotesize
\bibliographystyle{IEEEbib}
\bibliography{egbib}
}
\end{document}